\documentclass[11pt]{article}

\usepackage{acl}

\usepackage{times}
\usepackage{latexsym}

\usepackage[T1]{fontenc}

\usepackage[utf8]{inputenc}

\usepackage{microtype}

\usepackage{inconsolata}

\usepackage{graphicx}
\usepackage{amsmath}
\usepackage{amssymb}
\usepackage{bbm}
\usepackage{booktabs}
\usepackage{arydshln}
\usepackage{multirow}
\usepackage[most]{tcolorbox}

\title{ConsensusBench: Benchmark of Consensus Nodes for LLM Reasoning via Outcome Reward Densifying }

\author{
 \textbf{Shi-Qi Yan},
 \textbf{Chao-Hong Tan},
 \textbf{Qian Chen},
 \textbf{Wen Wang},
 \textbf{Xiangang Li},
 \textbf{Zhen-Hua Ling}
\\
\\
 Alibaba Token Hub, Alibaba Group
\\
 \small{
   sqyan01@mail.ustc.edu.cn, zhling@ustc.edu.cn
 }
}

\newcommand{\DATANAME}{\textsc{ConsensusBench}}

\begin{document}
\maketitle
\begin{abstract}
Reinforcement learning (RL) has become one of the primary paradigms for reasoning enhancement of large language models (LLMs). In particular, Group Relative Policy Optimization (GRPO) and related algorithms have demonstrated strong performance with outcome-level rewards. However, these methods depend solely on the final answer, without feedback regarding which intermediate steps contribute to success or failure. As task complexity and reasoning trajectory length increase, such sparse final-answer rewards become increasingly insufficient. To address this limitation, we introduce \DATANAME{}, a novel dataset designed to provide rule-based process-level signals. We posit that a correct final answer relies on a small set of intermediate conclusions throughout the reasoning process, which can be seen as a verifiable sub-outcome.
We identify these sub-outcomes by filtering correct trajectories from $N$ rollouts and clustering semantically equivalent intermediate statements. We call these clustered statements as \emph{Consensus Nodes}.
By integrating a rule-based process reward derived from these nodes into GRPO-style algorithms, we develop a new reinforcement learning signal named \emph{ConsensusPR}. It directly reduces the reward sparsity of outcome reward across long reasoning trajectories. 
To facilitate systematic process-level evaluation, we introduce three metrics to our benchmark: Final Answer Accuracy (Acc), Node Coverage Rate (NCR), and Tokens per Node (TPN). Experiments across AIME~2024, AIME~2025, GSM8K, MATH-500, and our \DATANAME{} demonstrate that the proposed method consistently surpasses GRPO-style approaches, highlighting the practical value of consensus nodes in guiding reasoning.

\end{abstract}

\section{Introduction}
\label{sec:intro}

\begin{figure*}[t]
\centering
    \vspace{-4mm}
    \includegraphics[width=1.0\textwidth]{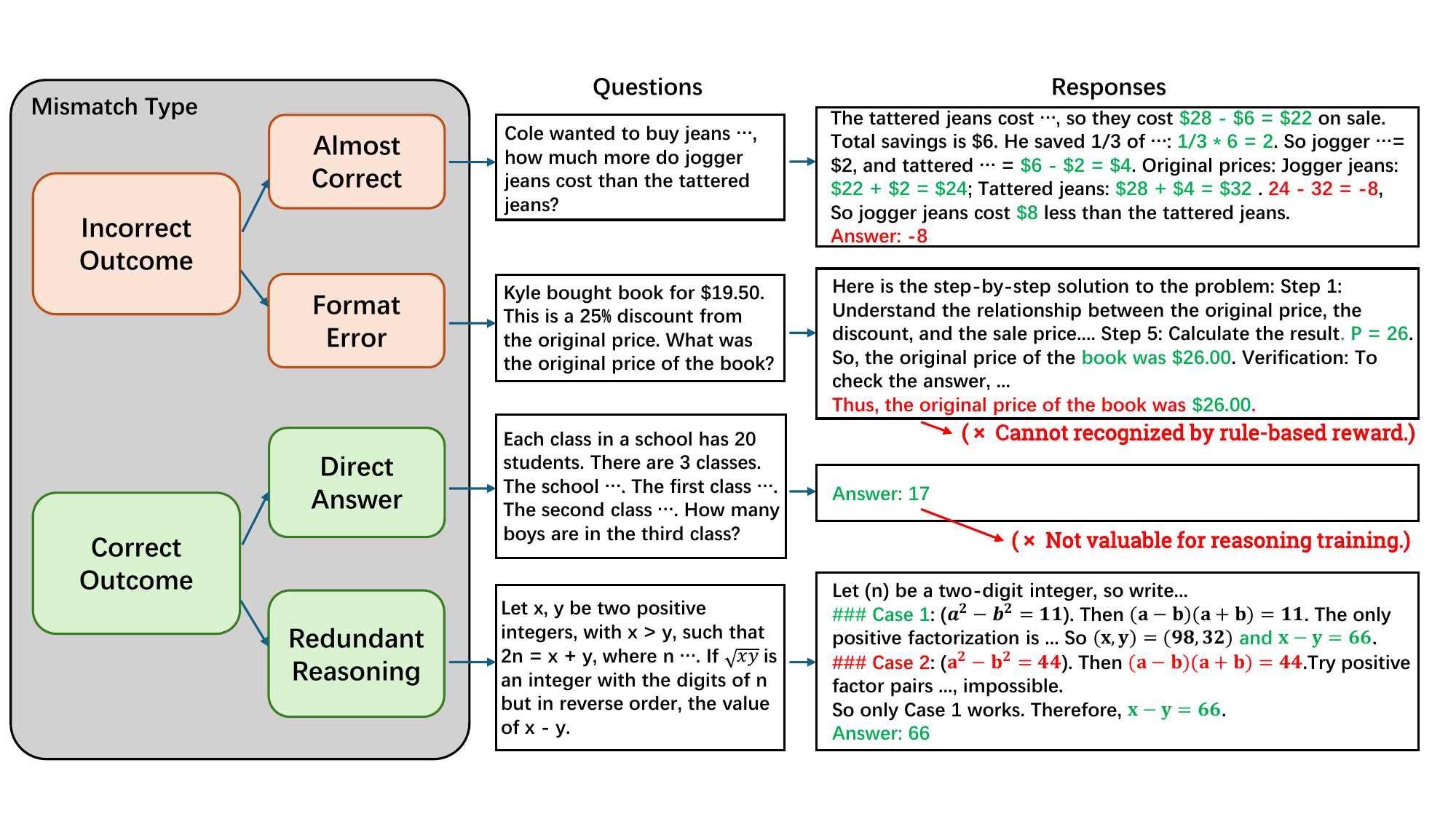}
    \vspace{-13mm}
    \caption{
    A case study of four main types of mismatch in outcome-only reward. The \textcolor{green}{green} spans in the responses indicate the correct reasoning content, while the \textcolor{red}{red} ones in the responses indicate the incorrect content.
    }
    \label{fig:case_study}
    \vspace{-4mm}
\end{figure*}

Recent advances in reinforcement learning (RL) have demonstrated significant capabilities for enhancing the reasoning of LLMs~\citep{DBLP:journals/tmlr/WeiTBRZBYBZMCHVLDF22, DBLP:conf/nips/Ouyang0JAWMZASR22}. Recent methods, including Group Relative Policy Optimization (GRPO)~\citep{DBLP:journals/corr/abs-2402-03300} and other Reinforcement Learning with Verifiable Rewards (RLVR)~\citep{DBLP:journals/corr/abs-2411-15124}, have reached a consensus that the outcome reward signal from the binary correctness of the final answer can be effectively used to drive policy optimization. Outcome rewards are readily verifiable without external human feedback, exhibiting low noise, high scalability, and strong resistance to reward hacking~\citep{DBLP:journals/corr/abs-2503-07572}. These properties have established the dominant role of outcome rewards in recent RL research~\citep{DBLP:journals/corr/abs-1811-07871}.

However, as reasoning chains expand to thousands of tokens for increasingly complex tasks, outcome-only supervision exhibits clear limitations~\citep{DBLP:journals/corr/abs-2604-20659}. 
A single scalar reward assigned to the entire trajectory provides virtually no information regarding intermediate progress, since uniform credit is distributed across all tokens. 
This formulation gives rise to two systematic challenges of reward mismatch. 
We conduct a case study and define four main types of reward mismatch as shown in Figure~\ref{fig:case_study}:
(1) Almost Correct: trajectories that reason correctly throughout almost the entire path but err only at the final step; 
(2) Format Error: trajectories that reason correctly but do not follow the output format of the answer in the instruction, which is not recognized by the answer matching rule;
(3) Direct Answer: trajectories directly answer the question without reasoning, which is not valuable for the model training;
(4) Redundant Thinking: trajectories that arrive at correct answers with tremendous redundant reasoning, which receive full credit like the correct reasoning.

Prior research has introduced the Process Reward Model (PRM)~\citep{DBLP:conf/iclr/LightmanKBEBLLS24,DBLP:conf/acl/Math-Shepherd}, which scores each intermediate reasoning step according to learned or human-annotated notions of quality. Although PRMs partially mitigate the credit assignment problem, they suffer from three severe challenges.
First, the considerable cost of collecting and annotating solutions presents a critical obstacle to scaling.
Second, training a dedicated process reward model introduces considerable computational overhead and impacts overall training efficiency.
Third, PRM methods require detailed solutions paired with corresponding quality annotations, leading to a semantically distinct reward signal in the training pipeline. The outcome reward concerns whether a trajectory reaches the correct destination, which is a binary and verifiable question, whereas the PRM focuses on the plausibility of every token, which is a subjective problem whose answer must be supplied by humans or by a learned policy, generalizes poorly out of distribution, and remains vulnerable to reward hacking when used to drive policy gradients~\citep{DBLP:conf/nips/ZelikmanWMG22}.

In this paper, we propose \DATANAME{}, a rule-based process reward dataset that densifies the outcome reward along the reasoning path, together with \emph{ConsensusPR}, a corresponding reinforcement learning reward.
We argue that correct final answers are generally composed of a sequence of intermediate outcomes through which the reasoning path frequently pass, such as subproblems resolved, constraints verified, cases partitioned, or lemmas applied.
Therefore, they are named as \emph{consensus nodes} in this paper, which constitute verifiable binary signals for an auxiliary reward that remains verifiable, low-noise, and objective-focused.
Correct rollouts, however verbally diverse, typically agree on the consistent intermediate checkpoints, thereby exhibiting process-level self-consistency~\citep{DBLP:conf/iclr/0002WSLCNCZ23}.
These consensus nodes are extracted automatically: $N$ rollouts per prompt are sampled, retaining the correct ones, and clustering the semantically equivalent intermediate conclusions.
We present the benchmark with three metrics, namely Final Answer Accuracy (Acc), Node Coverage Rate (NCR), and Tokens per Node (TPN), which together support a hierarchical view of how a trajectory behaves.
On the one hand, consensus nodes address the failure modes of binary outcome rewards, since all incorrect trajectories can be assigned a more appropriate score rather than a uniform zero.
On the other hand, the proposed reward integrates seamlessly with existing RLVR, while preserving their training stability.

Ultimately, our contributions are threefold. First, we provide a novel benchmark \DATANAME{} as well as its process reward annotation based on \emph{consensus nodes}. 
This balances the advantages of outcome-only and process-based reward: fine-grained, step-level signals can be obtained while preserving efficiency and generalization. 
Second, the reward derived from consensus nodes integrates seamlessly into existing GRPO and RLVR pipelines through \emph{ConsensusPR}, offering a more accurate and stable estimation. 
Third, Experiments on GRPO and DAPO demonstrate the performance and significance of our \emph{consensus nodes} and \DATANAME{}.

\section{Related Work}
\label{sec:related}
\paragraph{Outcome-based reinforcement learning for LLM reasoning.}
Trajectory-level outcome-based RL methods have become the dominant paradigm in LLM reasoning training.
Specifically, GRPO~\citep{DBLP:journals/corr/abs-2402-03300}, DAPO~\citep{DBLP:journals/corr/abs-2503-14476}, and GSPO~\citep{DBLP:journals/corr/abs-2507-18071} have produced strong gains on mathematics and code with a single binary signal of final-answer correctness, offering a lightweight and efficient algorithm compared with the critic-based ones like PPO~\citep{DBLP:journals/corr/SchulmanWDRK17}.
While they share a common structural limitation we target: every token in a large-scale reasoning trajectory receives the same scalar reward, leaving credit assignment essentially unsolved as chains grow longer. We preserve the outcome-reward objectives of this line and densify it along the reasoning in this paper.

\paragraph{Process reward models and benchmarks.}
A separate research line introduces critic-based models~\citep{DBLP:journals/corr/abs-2504-05118, DBLP:conf/icml/KazemnejadAPSRC25} or process reward models (PRMs)~\citep{DBLP:conf/iclr/LightmanKBEBLLS24,DBLP:conf/acl/Math-Shepherd} scoring individual intermediate steps.
The PRM800K~\citep{DBLP:conf/iclr/LightmanKBEBLLS24} dataset provides human-annotated labels for mathematical reasoning, training a learned scorer for verifier-guided decoding or as a step-level signal during policy optimization.
Subsequent work reduces annotation cost by replacing human labels with rollout- or search-based strategies, while a parallel series derives step rewards implicitly from outcome data~\citep{DBLP:conf/nips/ZelikmanWMG22}.
These approaches introduce a different reward-signal objective for training that judges reasoning quality rather than target achievement.
Tokens receive a high reward when they align with the annotated solution, regardless of their direct contribution to the final answer.
It leads to fragility: PRMs are expensive to train, often generalize poorly outside their step-label distribution, and are themselves susceptible to reward hacking when used as a policy-gradient signal.
Considering the diverse styles of validated solutions to a question, it is crucial for PRM to overcome the out-of-distribution challenge by training on a specific and well-annotated solution.
More details of the existing studies about PRMs are available in Appendix~\ref{app:related_work}
In contrast, rather than scoring step quality, we decompose the existing outcome reward into a sequence of intermediate, equally verifiable sub-outcomes, leaving every reward term's objective unchanged.

\section{\DATANAME{}}
\label{sec:bench}
\vspace{-2mm}

In this paper, we propose \DATANAME{}, a benchmark that contains sub-outcomes for process reward. 
Every sub-outcome is a verifiable target that correct solutions frequently reach, rather than a stylistic preference inherited from a single canonical path. 
Our approach integrates the strengths of outcome and process rewards, maintaining lightweight computation and dense supervision while circumventing the limitations inherent to each paradigm. By leveraging verifiable sub-outcomes, the method delivers fine-grained process rewards that outcome rewards cannot provide. Furthermore, because it eliminates the need for explicit solution annotation, it surpasses conventional process rewards in both generalization capability and data efficiency.
To the best of our knowledge, this is the first work to propose and construct a dataset with comparable objectives.
This section outlines \DATANAME{}'s design, source data, data construction pipeline, statistics, supported metrics, and the evaluation results on some main LLMs.

\vspace{-2mm}
\subsection{Design Principle}
\label{sec:bench-principle}
\vspace{-2mm}

\DATANAME{} is built on the principle that consensus nodes across correct rollouts serve as high-consensus sub-outcomes, while all other content remains unannotated. 
Existing process reward benchmarks generally provide step-level labels along a single canonical solution path, implicitly enforcing a specific reasoning style. We operate under the premise that multiple valid reasoning trajectories exist for a single problem. 
Annotating only a single path conflates correctness with the annotator's preferred style. Consequently, any different yet valid trajectory can be incorrectly treated, despite generating logically sound reasoning.

In contrast to prior benchmarks, we annotate only a set of consensus nodes in the reasoning process that are frequently reached by valid solutions. This annotation strategy yields three direct benefits. 
First, it enhances generalization by allowing multiple valid solution paths to satisfy the same set of checkpoints, which ensures that the model avoids penalties for selecting one valid trajectory over another. 
Second, consensus nodes mitigate the potential mismatch inherent in rule-based outcome rewards, as the count of covered nodes yields more calibrated rewards for reasoning trajectories. 
Finally, it first establishes a novel optimization target that encourages the model to maintain reasoning validity while pursuing outcome correctness.

\subsection{Data Construction}
\label{sec:bench-pipeline}

\DATANAME{} draws its problems from five public reasoning datasets: MATH (a curated subset)~\citep{DBLP:conf/nips/HendrycksBKABTS21}, GSM8K~\citep{DBLP:journals/corr/abs-2110-14168}, AIME-2024~\citep{aime}, and AIME-2025~\citep{aime} for the test set, and DAPO-Math-17K~\citep{DBLP:journals/corr/abs-2503-14476} for the training set.
For each problem $q$ in the source pool, we apply the three-stage extraction pipeline, instantiated as follows.

\vspace{-2mm}
\paragraph{Stage 1: Rollout sampling and correctness filtering.}
We sample $N = 5$ correct rollouts from open models at temperature $T = 2.0$ (we use GPT-5.2~\citep{gpt-5.2}, GPT-5-mini~\citep{gpt-5}, Qwen3.5-32B~\citep{qwen3}, Minimax-M2~\citep{minimax-m2}, and Gemini-2.5-pro~\citep{Gemini-2.5} as the rollout models for the different styles of trajectories; results reported below use the union over the these models).
Rollouts are filtered for final-answer correctness using the automatic verifier of each source dataset.
Considering that some questions are too difficult to obtain enough $N$ rollouts, problems with fewer than $N_{\min} = 3$ correct rollouts after sampling are excluded, this guarantees that each retained problem has a statistically meaningful consensus signal. 

\vspace{-2mm}
\paragraph{Stage 2: Node Extraction and Expression Expansion.}
In this stage, an open model (GPT-5.2~\citep{gpt-5.2} in this paper) is driven to extract consensus nodes among the filtered rollouts.
By prompting the model, the concepts that consistently appear in the rollouts are extracted and reformed into claim nodes. Since the math domain is mainly focused on in this paper, these nodes are often built based on a formula or proposition.
Subsequently, for a robust verification, the nodes are rewritten with different forms of expressions.

\vspace{-2mm}
\paragraph{Stage 3: Node refinement.}
To enhance the verifiability and supervisory value of nodes, we further refine them by removing trivial clusters that merely restate the problem or review definitions, which typically appear at the beginning and end of trajectories. 
Conversely, we prioritize retaining nodes with numbers and formulas, as they offer both high verifiability and supervisory value.

The output of this pipeline is, for each problem $q$, an ordered set ${Node}(q) = \{node_1, \ldots, node_k\}$ of consensus nodes, each accompanied by its canonical label and its type. 
Hyperparameters $N$, $T$, $|Node|_{\min}$, and $|Node|_{\max}$ are kept fixed across all source datasets to ensure that the benchmark is constructed under a single, transparent strategy.

\subsection{Dataset Statistics}
\label{sec:bench-stats}

\DATANAME{} contains \textbf{5,577} problems drawn from the five sources in total as described above, with an average of \textbf{6.80} consensus nodes per problem. 
Following established practices in prior work, we designate the first four datasets as the test split, which comprises 1,707 samples with an average of 4.88 nodes. 
While the DAPO-MATH-17K dataset serves as the training split and contains 3,870 samples averaging 7.64 nodes.
We report the complete statistics in Table~\ref{tab:bench-composition}. 

\begin{table}[t]
\centering
\caption{Composition and statistics of \DATANAME{}. }
\vspace{-4mm}
\label{tab:bench-composition}
\begin{tabular}{lcc}
\toprule
Source            & \#Problems & Avg.\ $N(k)$\\
\midrule
MATH-500                    & 441   & 5.36 \\
GSM8K                       & 1212  & 4.53 \\
AIME-2024                   & 30    & 8.50 \\
AIME-2025                   & 24    & 7.88 \\
DAPO-MATH-17K               & 3870  & 7.64 \\
\midrule
\textbf{Total} (Test)       & 1707  & 4.88 \\
\textbf{Total} (Train)      & 3870  & 7.64 \\
\bottomrule
\end{tabular}
\vspace{-5mm}
\end{table}

\subsection{Evaluation Metrics}
\label{sec:bench-metrics}

\DATANAME{} supports the standard final-answer accuracy (Acc) used by its source datasets. 
Still, we also present two more process-level metrics: node coverage rate (NCR) and tokens per node (TPN), that diagnose how a model's intermediate reasoning behaves. 
Let $\hat{o}$ denote a model's rollout and let ${Node}(q) = \{n_1, \ldots, n_k\}$ be the annotated consensus nodes of the problem $q$.

\paragraph{Node Coverage Rate (NCR).}
The fraction of annotated nodes that the model's rollout attains, judged by matching the existence of the nodes in the content of the rollout:
\begin{equation}
\label{eq:ncr}
\mathrm{NCR}(\hat{o}, q)
= \frac{\sum_{n \in {Node}(q)}\mathbbm{1}\bigl[n \text{ is matched in } \hat{o}\bigr]}{|{Node}(q)|}.
\end{equation}
NCR is the direct reference metric of final-answer accuracy lifted to the sub-outcome level.

\paragraph{Tokens per Node (TPN).}
The average number of tokens that a rollout requires to reach the next consensus node, used to evaluate the reasoning efficiency of a rollout:
\begin{equation}
\label{eq:rs}
\mathrm{TPN}(\hat{o}, q)
= \frac{\left | \hat{o} \right |}
       {\sum_{n \in {Node}(q)}\mathbbm{1}\bigl[n \text{ is matched in } \hat{o}\bigr]}.
\end{equation}
Very low TPN indicates that the rollout is efficient without much redundant thinking; very high TPN may indicate a high fraction of invalid reasoning or unnecessary trials.
TPN is therefore intended as an auxiliary complement to NCR and Acc rather than as a direct quality score.

\subsection{Consensus Nodes Validation}
To validate the generalizability and robustness of our proposed consensus nodes, we examine their effectiveness across diverse out-of-distribution styles, their consistency with final answer rewards, and their agreement with human annotations in Appendix~\ref{app:human_validation}.
We evaluate \DATANAME{} with an independent model (DeepSeek-V4-pro~\citep{deepseek-v4} in this paper), which does not participate in data construction or annotation, and compare its NCR scores against existing in-domain baselines. The results in Table~\ref{tab:leaderboard} demonstrate that the NCR scores maintain a stable interval, revealing the generalizability of our consensus nodes across different models with various response styles. 
Subsequently, the consistency between model Acc and NCR scores is quantified, as demonstrated in Table~\ref{tab:consistency-acc-ncr}. Three main metrics are involved, including the difference of the average NCR scores between correct and incorrect samples $D_\text{NCR}$, AUROC~\citep{AUROC}, and AUCPR~\citep{AUCPR}. Results on independent models exhibit no significant performance degradation, and the consistency between NCR and Acc remains robust.
Detailed experiment settings and results are available in Appendix~\ref{app:results}.
Multi-dimensional experiments in this section further prove the effectiveness and generalizability of our \DATANAME{}.

\vspace{-2mm}
\section{ConsensusPR}
\label{sec:method}
\vspace{-2mm}

Since \DATANAME{} provides verifiable consensus nodes that act as process-level sub-outcomes, these nodes enable the deployment of verifiable process rewards that supply consistent intermediate feedback.
In contrast to previous process reward models, consensus nodes operate as necessary checkpoints that a rollout frequently passes to reach a correct final answer.
Rewards assigned at these nodes thus constitute a temporally densified decomposition of the outcome reward.
Notably, this design guarantees that the optimization direction of our process rewards aligns closely with that of the outcome reward, thereby avoid the generalization limitations and out-of-distribution vulnerabilities that typically stem from misaligned objectives in PRMs.
To further assess the advantage of our process supervision over outcome-only feedback, we design three different reinforcement learning rewards. 
Subsequently, we conduct a simple but useful training by seamlessly integrating these rewards into a GRPO-style framework and developing a corresponding reinforcement learning training paradigm.

\begin{table}[t]
\centering
\caption{Consistency metrics between Acc and NCR across models }
\vspace{-4mm}
\label{tab:consistency-acc-ncr}
\begin{tabular}{lccc}
\toprule
Model            & $D_\text{NCR}$ & AUROC & AUCPR\\
\midrule
Gemini-2.5-pro   & 30.5 & 76.9 & 96.61 \\
MiniMax-M2       & 33.9 & 74.1 & 96.65 \\
GPT-5.2          & 18.0 & 65.5 & 96.58 \\
DeepSeek-V4-pro  & 16.7 & 63.0 & 96.38 \\
\bottomrule
\end{tabular}
\vspace{-5mm}
\end{table}

\subsection{Training Overall}
\vspace{-2mm}
We directly integrate our process reward into the GRPO-style training process. 
To validate the effectiveness of consensus nodes as a process reward signal and demonstrate their natural alignment with the outcome reward. 
We deliberately avoid staged reward scheduling or complex training mechanisms, adopting instead a minimal coupling strategy that preserves the original reward structure. 
Specifically, we integrate node-level signals as an additive term directly into the standard GRPO scalar reward. 
This design follows from our conceptual premise that consensus nodes represent a temporal extension of the final outcome reward with identical semantics and verifiability.
Introducing auxiliary components during training would implicitly assign distinct optimization roles to the two signals, contradicting our core hypothesis. 

\subsection{Node-Aware Reward Design}
\label{sec:reward_design}
The core of our method lies in converting consensus nodes into a training-time reward signal that remains semantically identical to the outcome reward while providing fine-grained temporal supervision.

\subsubsection{Node Matching at Training Time}
During RL training, for a newly sampled rollout $r_{\text{train}} = [x_1, \dots, x_T]$, we evaluate whether each consensus node $n_k \in \mathcal{N}$ has been achieved. 
In this experiment, we introduce a rule-based soft matching strategy to compute process rewards. 
Given that the extracted nodes primarily consist of formulas, equations, and propositions carrying unique verifiable mathematical values, we formalize these nodes through rule-based text normalization and match them against the formalized rollouts.
This alignment yields a fast and stable reward computation mechanism. 
The resulting matching metric, defined as the Node Coverage Ratio (NCR), serves directly as the process reward score.

\subsubsection{Reward Formulation}
Considering the diverse integration strategies for outcome and process rewards, we design three distinct reward formulations to systematically evaluate the impact of injecting varying proportions of process rewards into GRPO-style reinforcement learning algorithms.
We propose three forms of reward integration strategies as follows.

\paragraph{Linear Integration.}
The most natural method is to linearly combine process and outcome rewards. 
We introduce a weighting hyperparameter $\alpha$ (0.5 by default in this work) to balance the final answer accuracy (Acc) and the node coverage ratio (NCR), producing a composite reward score. 
Within this scheme, the final answer operates as a special node with higher weight, demonstrating the inherent compatibility between our process-level node rewards and the outcome signal.
The reward can be formulated as:

\begin{equation}
    r_{\text{linear}} = \alpha \cdot \text{Acc} + (1-\alpha) \cdot \text{NCR},
    \label{eq:reward-linear}
\end{equation}
where $r_{\text{outcome}} \in \{0, 1\}$ is the standard terminal outcome reward (coarsest granularity), $\text{NCR}$ (Node Coverage Rate) $ \in [0, 1]$ measures the proportion of achieved consensus nodes, $\alpha$ is a scalar weight balancing process and outcome reward.

\paragraph{Outcome-Priority Integration.}
We introduce an outcome-priority reward mechanism. 
Nodes appearing later in a sequence are inherently more difficult to match due to the temporal causality of rollouts. 
The successful alignment of subsequent nodes typically depends on the accuracy of preceding reasoning steps, whereas earlier nodes remain independent of later content. 
Consequently, treating the final answer merely as a larger node may encourage the model to procrastinate the final answer and instead exploit the reward signal by generating additional intermediate nodes. 
To mitigate this issue, we prioritize the outcome reward. 
The model receives the maximum score when it generates a correct final answer, and the evaluation falls back to nodes matching only when the final answer is incorrect.
The reward can be formulated as:

\begin{equation}
    r_{\text{max}} = \max \left \{ \text{Acc}, \text{NCR} \right \}.
    \label{eq:reward-max}
\end{equation}

\paragraph{Curriculum Learning}
Finally, we also implement a curriculum learning strategy.
As noted above, aligning with consensus nodes presents a simpler task for the model, whereas deriving the correct final answer requires more complex reasoning.
We therefore structure a two-stage curriculum. 
During the first stage, the model receives exclusively process rewards based on node matching, focusing entirely on the quality and accuracy of intermediate reasoning steps without any requirement to produce a final answer.
We switch the curriculum after the first epoch, where the models can focus on the process (high NCR) but do not know how to answer the question (low Acc). The second stage switches exclusively to outcome rewards, directing the full optimization capacity toward deriving the correct final answer.

\section{Experiments}
\label{sec:experiments}

\begin{table*}[t]
\centering
\caption{Main results across reasoning benchmarks (pass@1, pass@16, NCR, \%). Bold denotes best, underline denotes second best. All models initialized from the same base checkpoints.}
\label{tab:main_results}
\begin{tabular}{lcccccc}
\toprule
 & \textbf{AIME 24} & \textbf{AIME 25} & \textbf{GSM8K} & \textbf{MATH-500} & \multicolumn{2}{c}{\textbf{\DATANAME{}}} \\
\textbf{Method} & Acc@16 & Acc@16 & Acc@1 & Acc@1 & Acc & NCR \\
\midrule
\multicolumn{7}{c}{\emph{Qwen3-4B-Base}} \\
\midrule
Vanilla GRPO & 23.3 & 16.7 & 92.1 & 54.4 & 83.0 & 75.5 \\
 + $\text{ConsensusPR}_\text{linear}$ & 33.3 & 20.0 & \textbf{92.8} & 59.0 & \textbf{84.7} & 76.8 \\
 + $\text{ConsensusPR}_\text{max}$ & \textbf{36.7} & 23.3 & 92.6 & 59.2 & 84.5 & \textbf{77.4} \\
 + $\text{ConsensusPR}_\text{CL}$ & 33.3 & \textbf{26.7} & 83.0 & \textbf{62.4} & 78.2 & 76.5 \\
\midrule
Vanilla DAPO & 26.7 & 26.7 & 92.3 & 58.4 & 84.1 & 77.4 \\
 + $\text{ConsensusPR}_\text{linear}$ & 33.3 & \textbf{36.7} & 92.8 & 60.6 & 84.9 & \textbf{78.4} \\
 + $\text{ConsensusPR}_\text{max}$ & \textbf{40.0} & 30.0 & 92.9 & 59.6 & 84.9 & 77.6 \\
 + $\text{ConsensusPR}_\text{CL}$ & \textbf{40.0} & 26.7 & \textbf{93.3} & \textbf{66.6} & \textbf{87.2} & 76.7 \\
\midrule
\multicolumn{7}{c}{\emph{Qwen3-8B-Base}} \\
\midrule
Vanilla GRPO & 40.0 & 30.0 & 79.1 & 76.0 & 81.7 & 73.7 \\
 + $\text{ConsensusPR}_\text{linear}$ & 40.0 & 40.0 & 84.0 & 73.2 & 85.2 & 74.7 \\
 + $\text{ConsensusPR}_\text{max}$ & {40.0} & \textbf{43.3} & {88.1} & {74.4} & {87.6} & {74.7} \\
 + $\text{ConsensusPR}_\text{CL}$ & \textbf{46.7} & 36.7 & \textbf{88.7} & \textbf{76.6} & \textbf{89.2} & \textbf{75.9} \\
\midrule
Vanilla DAPO & 43.3 & 36.7 & 87.1 & 70.6 & 85.1 & 75.9 \\
 + $\text{ConsensusPR}_\text{linear}$ & 43.3 & 40.0 & 87.0 & 70.8 & 85.2 & \textbf{76.7} \\
 + $\text{ConsensusPR}_\text{max}$ & 46.7 & 43.3 & 90.6 & 76.4 & 88.6 & 73.6 \\
 + $\text{ConsensusPR}_\text{CL}$ & \textbf{53.3} & \textbf{53.3} & \textbf{91.2} & \textbf{77.6} & \textbf{89.7} & 75.8 \\
\bottomrule
\end{tabular}
\vspace{-2mm}
\end{table*}

We empirically validate our core hypothesis: densifying outcome rewards via consensus nodes resolves credit assignment sparsity in long-chain reasoning without introducing heterogeneous rewards or human step-level annotations. Our experiments address three questions: (1) Does ConsensusPR consistently outperform outcome-only baselines? (2) How do the performance gains achieved by ConsensusPR correlate with the inherent difficulty of the task and the length of the reasoning?  (3) Do gains persist as the model scale varies?

\subsection{Experimental Setup}
\paragraph{Models and Framework.}
We evaluate three open-weight base models spanning different scales: Qwen3-1.7B-Base, Qwen3-4B-Base, and Qwen3-8B-Base~\citep{qwen3}. 
All experiments are conducted within the GRPO and DAPO framework using the \texttt{verl} libraries. 
The RL optimization pipeline remains unmodified; only the reward computation module is replaced with our node-aware formulation (\S\ref{sec:reward_design}).

\paragraph{Training Details.} Models are trained on a training set of \DATANAME{}. 
For each prompt, we sample $N=8$ rollouts during the training phase.
We set the hyperparameter $\alpha=0.5$ in $\text{ConsensusPR}_\text{linear}$, in case that models are rewarded to focus on intermediate nodes and refuse to give the final answer.
Full hyperparameters, hardware configurations, and reproducibility details are provided in Appendix~\ref{app:hyperparams}.

\paragraph{Evaluation Benchmarks.} We report performance on four established reasoning suites: AIME 2024, AIME 2025, GSM8K, and MATH-500, individually as well as the test set of our \DATANAME{} (\S\ref{sec:bench}). 
Additionally, node-level metrics is utilized during evaluation on \DATANAME{}: Node Coverage Rate (NCR), and Tokens per Node (TPN). 
Pass@16 results are reported on AIME, and @1 are reported on the other benchmarks for the final answer accuracy (Acc).

\subsection{Baselines}
Since our original motivation is to evaluate the significance and performance of the proposed consensus nodes, we simply compare against the outcome-only baselines without process supervision: 
(1) \textbf{Vanilla GRPO:}  a value-free RLVR that optimizes language model policies by sampling multiple responses per prompt, computing advantages relative to the group's average reward without a separate critic network. 
(2) \textbf{Vanilla DAPO:} DAPO enhances GRPO by replacing its static advantage with a dynamic, variance-adaptive advantage normalization that stabilizes training and improves sample efficiency while preserving GRPO's value-free architecture. 
All baselines are trained under identical data budgets, compute constraints, and optimization schedules to ensure fair comparison.

\subsection{Main Results}
We conduct further analysis and respond to the three research questions above in this section:

\paragraph{RQ1: Does ConsensusPR consistently outperform outcome-only baselines?} 
Table~\ref{tab:main_results} summarizes performance across all benchmarks. 
All three forms of ConsensusPR achieves consistent gains over vanilla GRPO and vanilla DAPO.
Specifically, $\text{ConsensusPR}_\text{CL}$ drives consistent gains on complex reasoning tasks. When integrated with DAPO on the Qwen3-4B-Base checkpoint, it elevates MATH-500 accuracy from 58.4\% to 66.6\% and improves AIME 24 scores to 40.0\%, respectively. This trend persists on the larger Qwen3-8B-Base model, where the same configuration achieves peak results on GSM8K (91.2\%), MATH-500 (77.6\%), and the \DATANAME{} accuracy metric (89.7\%). The superiority of the CL approach suggests the significance of the consensus nodes searching in the applications, effectively stabilizing policy optimization on multi-step reasoning trajectories.

Additionally, different reward aggregation mechanisms exhibit different optimization biases. While $\text{ConsensusPR}_\text{CL}$ maximizes absolute accuracy on challenging benchmarks, the linear and max ones frequently achieve higher nodes covering scores. For instance, on Qwen3-4B-Base with DAPO, $\text{ConsensusPR}_\text{linear}$ reached the highest NCR metric (78.4\%) despite trailing the CL approach by 2.3\% on the \DATANAME{} accuracy. Similarly, under the GRPO framework, $\text{ConsensusPR}_\text{max}$ secures the top NCR score (77.4\%) while the CL variant experiences a notable drop in GSM8K performance (83.0\% versus 92.1\% for vanilla GRPO).

\begin{table}[t]
\centering
\caption{Tokens per node (TPN) scores across models.}
\label{tab:tpn_results}
\resizebox{\columnwidth}{!}{
\begin{tabular}{lcc}
\toprule
 & \textbf{\small{Qwen3-4B-Base}} & \textbf{\small{Qwen3-8B-Base}} \\
 \textbf{Methods} & TPN & TPN \\
\midrule
vanilla GRPO & \textbf{433.95} & \textbf{1527.35} \\
+ linear     & 744.13 & 1786.49 \\
+ max        & 564.43 & 1792.43 \\
+ CL         & 562.92 & 1529.14 \\
\midrule
vanilla DAPO & \textbf{320.47} & 1349.12 \\
+ linear     & 726.26 & \textbf{1048.65} \\
+ max        & 446.90 & 1373.83 \\
+ CL         & 408.13 & 1342.03 \\
\bottomrule
\end{tabular}
}
\end{table}

\paragraph{RQ2: How do the performance gains correlate with the inherent difficulty of the task?} 
Our experiments demonstrate that integrating process reward signals derived from the nodes yields consistent improvements across multiple benchmarks. The performance gains from ConsensusPR also scale with task difficulty. Improvements remain marginal on the straightforward GSM8K dataset, while the method delivers substantial gains on the more challenging MATH dataset and the highly complex AIME benchmark. 
One potential reason that leads to this pattern is the similar difficulty distribution between these benchmarks and the training data.
To quantify difficulty and provide a more intuitive presentation, we divide the MATH-500 dataset into easy, medium, and hard groups according to its difficulty labels. The experimental results in Table~\ref{tab:difficulty_results} yield consistent conclusions.
Concurrently, increased task difficulty substantially extends the reasoning trajectory, a behavior that exposes the limitations of traditional outcome rewards on long reasoning chains and confirms the effectiveness of our verifiable process reward mechanism. 

\begin{table}[t]
\centering
\caption{Difficulty Analysis on Qwen3-4B-Base.}
\label{tab:difficulty_results}
\resizebox{\columnwidth}{!}{
\begin{tabular}{lccc}
\toprule
 \textbf{Methods} & \textbf{Easy} & \textbf{Medium} & \textbf{Hard} \\
\midrule
\multicolumn{4}{c}{\textbf{Acc}} \\
\hdashline
DAPO        & 76.7 & 62.8 & 41.8 \\
+ linear    & 76.7(+0.0\%) & 64.4(+2.5\%) & \textbf{46.3(+10.8\%)} \\
+ max       & 79.1(+1.3\%) & 64.1(+2.1\%) & 42.5(+1.7\%) \\
+ CL        & 83.7(+9.1\%) & 71.5(+13.8\%) & \textbf{49.3(+17.9\%)} \\
\midrule
\multicolumn{4}{c}{\textbf{Response Length}} \\
\hdashline
DAPO        & 357.3 & 1244.8 & 1760.8 \\
+ linear    & 2219.0 & 3336.6 & 5755.2 \\
+ max       & 736.4 & 1678.0 & 2427.8 \\
+ CL        & 501.5 & 1682.4 & 2659.0 \\
\bottomrule
\end{tabular}
}
\end{table}

\paragraph{RQ3: Do gains persist as the model scale varies?} 
We evaluate ConsensusPR on 1.7B, 4B, and 8B parameter models from the Qwen3 family. 
The results of Qwen3-1.7B are available in Appendix~\ref{app:results}
DAPO demonstrates stronger compatibility with the proposed modules than GRPO, particularly on the 4B scale, when combining with $\text{ConsensusPR}_\text{CL}$. 
Scaling to the 8B further amplifies gains on structured mathematical tasks, with rising accuracy by 11.0\% on MATH-500, and 2.5\% on our \DATANAME{} relative to the 4B equivalent under identical training configurations. 
While the lower performance of 8B models on GSM8K than the 4B ones is observed as well. We attribute this degradation to the constrained generalization capacity of models at this scale.
The model tends to prioritize patterns that align with the difficulty distribution of the training data (DAPO-Math-17k), which inevitably weakens its fitting capability on simpler benchmarks such as GSM8K.

\subsection{Efficiency Analysis}
\paragraph{Tokens per Node (TPN).}
Table~\ref{tab:tpn_results} demonstrates the TPN scores across the different models. 
Vanilla GRPO and DAPO generally produce the shortest reasoning lengths, while the three reward methods, increase the reasoning length to varying degrees. The results is intuitive and reasonable. Although our minimal coupling strategy directly integrates process rewards into the outcome reward, which limits the model capacity to assess reasoning completeness, this behavior nevertheless confirms that the proposed reward mechanism effectively reinforces stepwise reasoning.

\begin{figure*}[t]
\centering
    \includegraphics[width=1.0\textwidth]{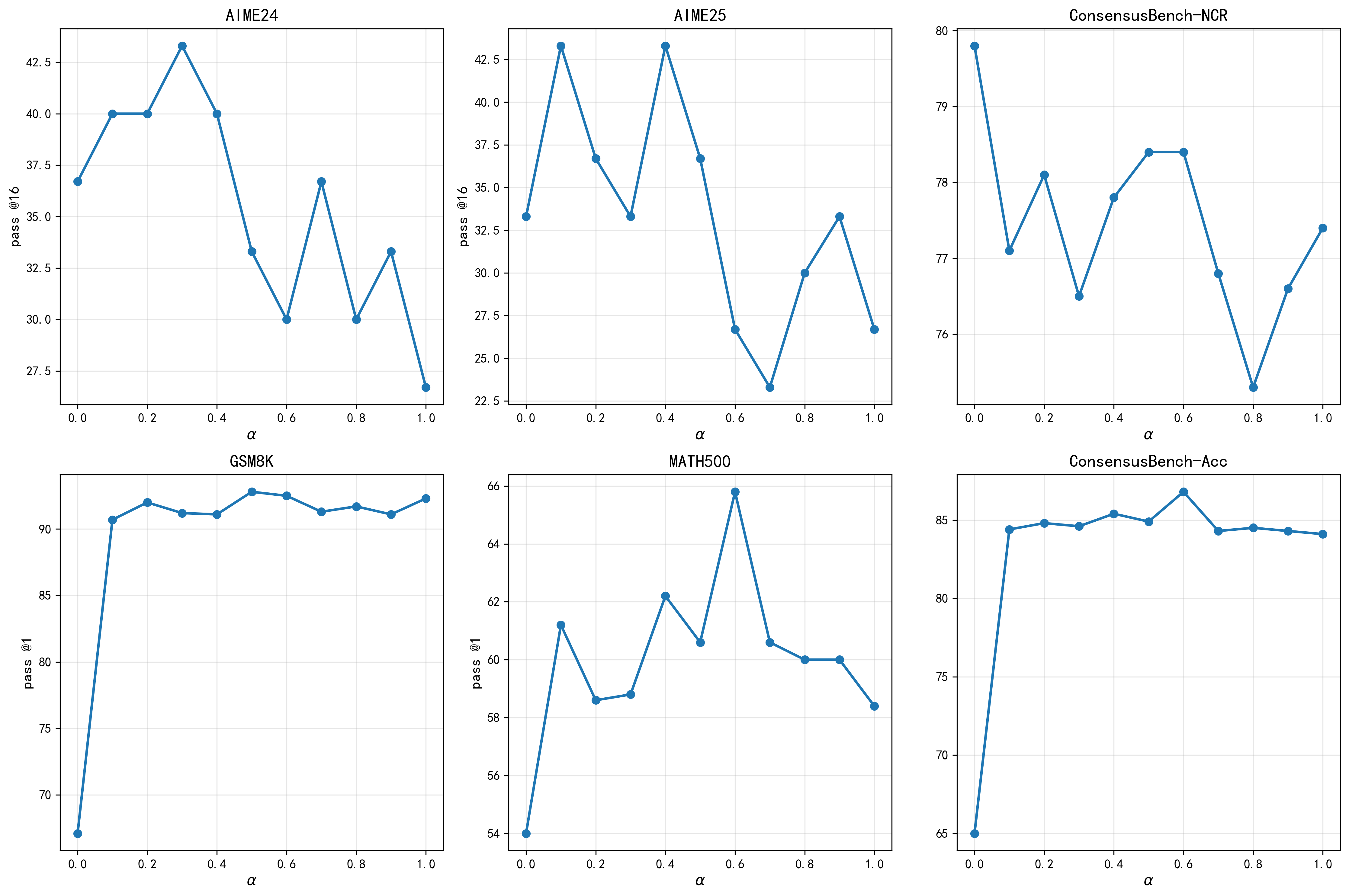}
    \caption{
    Performance curves of Qwen3-4B-Base across metrics when trained with different values of the hyperparameter $\alpha$. More detailed data are provided in Appendix~\ref{app:hyperpara}.
    }
    \label{fig:hyperpara_study}
\end{figure*}

\subsection{Hyperparameter Sensitivity Studies}
Since a hyperparameter $\alpha$ is introduced during the linear integration method $\text{ConsensusPR}_\text{linear}$, we specifically study the model's sensitivity to it.
As shown in Figure~\ref{fig:hyperpara_study}, the overall performance on ConsensusBench is relatively insensitive to the exact value of $\alpha$. As long as $\alpha$. avoids the degenerate extremes (0.0 or 1.0), the model consistently maintains high performance, demonstrating the robustness of our consensus framework.
On the other hand, We also observe an interesting divergence based on task complexity. A smaller $\alpha$ (which inherently assigns a higher weight to the process/node reward) yields superior performance on highly reasoning tasks (AIME and NCR). Conversely, a larger $\alpha$ around 0.6 (favoring the outcome) leads to better performance on simpler, more straightforward tasks.
More detailed resutls are available in Appendix~\ref{app:hyperpara}

\vspace{-2mm}
\section{Conclusion}
\label{sec:conclusion}
\vspace{-2mm}

In this paper, we have argued that outcome-only RLVR methods have challenges in long-chain LLM reasoning, where a sparse binary signal is limited for a fine-grained supervision.
Therefore, we propose and confirm the \emph{consensus nodes} in the LLM reasoning, which rests on a concept: any correct final answer to a complex question can be decomposed into a small set of intermediate conclusions that the trajectory frequently reached. 
Based on this principle, we collected consensus nodes by clustering reasoning consensus among correct trajectories in $N$ rollouts, and create a benchmark named \DATANAME{}.
\DATANAME{} provides the annotations of the verifiable consensus nodes that every valid solution generally traverse for process reward, and exposes three metrics (Acc, NCR, TPN) that diagnose reasoning under the same semantic discipline used during training.
Subsequent, \emph{ConsensusPR} is proposed to integrate the signal of verifiable process reward into previous outcome-only RLVR reward to validate the effectiveness of the consensus nodes.
Three main integration strategies are proposed and the experiments on various benchmarks consistently demonstrate the performance of our proposal.

\section*{Limitations}

In this work, we introduce a benchmark constructed based on the consensus nodes. 
To facilitate a direct comparison with outcome-reward methods such as GRPO and DAPO, we do not devise a standalone process-reward algorithm. 
Instead, we integrate process rewards directly into the outcome signal. Although empirical results validate the effectiveness of our \DATANAME{}, the current framework does not fully exploit the capabilities of consensus nodes. Additionally, since several closed-source models are utilized for data construction, the risks of the potential bias in the node extraction processing deserve focus as well. Subsequent research will further focus on addressing this limitation by developing dedicated process-reward formulations.

\bibliography{custom}

\clearpage
\appendix

\begin{table*}[]
\centering
\caption{Results across reasoning benchmarks (pass@1, pass@16, NCR, (\%), TPN($N_\text{tokens}$)) on Qwen3-1.7B-Base.}
\label{tab:1.7b-results}
\begin{tabular}{lccccccc}
\toprule
 & \textbf{AIME 24} & \textbf{AIME 25} & \textbf{GSM8K} & \textbf{MATH-500} & \multicolumn{3}{c}{\textbf{\DATANAME{}}} \\
\textbf{Method} & Acc@16 & Acc@16 & Acc@1 & Acc@1 & Acc & NCR & TPN \\
\midrule
\multicolumn{8}{c}{\emph{Qwen3-1.7B-Base}} \\
\midrule
Vanilla GRPO & 10.0 & 6.67 & \textbf{81.3} & 49.2 & 73.4 & 70.7 & 1,927 \\
 + $\text{ConsensusPR}_\text{linear}$ & \textbf{16.7} & \textbf{16.7} & 71.4 & 57.8 & 70.3 & \textbf{71.3} & 1,340 \\
 + $\text{ConsensusPR}_\text{max}$ & 13.3 & 10.0 & 78.1 & 59.0 & \textbf{74.7} & 71.1 & \multicolumn{1}{c}{1,265} \\
 + $\text{ConsensusPR}_\text{CL}$ & 6.7 & 6.7 & 75.2 & \textbf{59.4} & 73.2 & 69.9 & \multicolumn{1}{c}{677} \\
\midrule
Vanilla DAPO & 13.3 & 10.0 & 82.3 & 53.6 & 75.4 & 69.2 & 1,381 \\
 + $\text{ConsensusPR}_\text{linear}$ & 16.7 & \textbf{13.3} & 82.3 & 54.8 & 76.1 & \textbf{70.3} & 1,412 \\
 + $\text{ConsensusPR}_\text{max}$ & \textbf{26.7} & \textbf{13.3} & 82.6 & 55.2 & \textbf{76.7} & 70.2 & 1,239 \\
 + $\text{ConsensusPR}_\text{CL}$ & 23.3 & \textbf{13.3} & \textbf{83.2} & \textbf{57.0} & 73.0 & 69.5 & 1,824 \\
\bottomrule
\end{tabular}
\vspace{-5mm}
\end{table*}

\section{Ethics Statement}
We declare that AI was used solely for language polishing and revision in this paper, and NO factual content, such as data or citations, was generated by AI. 
All assets involved in this work are open-source and strictly comply with their respective license requirements. 
We will release the assets created in this paper, such as experimental data and code, which will be distributed under the CC-BY 4.0 license.

\section{Related Work}
\label{app:related_work}
\paragraph{Self-consistency and verifier-free methods.}
Previous studies observe the self-consistency where correct answers cluster across independent samples~\citep{DBLP:journals/corr/abs-2308-08998, DBLP:conf/icml/YuanPCLSXW24}.
These methods exploit consistency in the final-answer space.
Our work pushes this observation further: although correct rollouts diverge widely in surface form, they tend to converge on the same necessary intermediate conclusions, and these process-level consistencies can be harvested as a training signal.
Equivalently, we replace self-consistency's answer-level voting with intermediate-node voting, strictly preserving its binary, verifiable semantics.

\paragraph{Process-level reasoning benchmarks.}
Since most widely used reasoning benchmarks evaluate only the final answer~\citep{DBLP:journals/corr/abs-2110-14168,DBLP:conf/nips/HendrycksBKABTS21,DBLP:conf/acl/SuzgunSSGTCCLCZ23, DBLP:conf/iclr/JainHGLYZWSSS25}, a small body of process-aware benchmarks~\citep{DBLP:conf/iclr/LightmanKBEBLLS24,DBLP:conf/acl/ZhengZZLLYLZL25,DBLP:conf/nips/ZengLWLCDYXQZSL24} provides fine-grained annotations along a single canonical solution path, supporting stepwise correctness and error localization evaluation.
These resources are valuable, but their reliance on a single annotated path implicitly encodes a stylistic preference: a model solving the problem along a different but equally valid path is treated as out-of-distribution, even when its reasoning is sound.
\DATANAME{} takes the opposite stance, annotating high-consensus sub-outcomes that are frequently reached by diverse correct rollouts.
This preserves process-level evaluation's diagnostic value while accommodating the multiplicity of valid solution paths inherent to nontrivial reasoning.
To our knowledge, this is the first benchmark designed around high-consensus sub-outcomes rather than annotated step quality.

\section{Experiments Details}
\label{sec:appendix}

\subsection{Prompt}
We follow the previous DAPO-MATH-17K and utilize the same prompt in this paper for question answering to confirm that the LLMs can think carefully and answer in a unified form.
The feasibility and effectiveness of such CoT prompts have been extensively validated by prior research. 
\begin{tcolorbox}[colback=gray!5!white,colframe=gray!75!black,title=Prompt for Quesiton Answering]
  Solve the following math problem step by step. The last line of your response should be of the form Answer: \$Answer (without quotes) where \$Answer is the answer to the problem.\\
  
  \textsc{your question here}\\
  
  Remember to put your answer on its own line after "Answer:".
\end{tcolorbox}

\subsection{Training Details}
\label{app:hyperparams}
We run our experiments on a machine with 4 $\times$ 80 GB GPUs. To ensure experimental consistency and validity, we retain the default configurations from the verl examples for the majority of parameters.
We adjust only parameters such as \texttt{response\_length} and \texttt{batch\_size} to prevent out-of-memory errors during training, and we maintain these adjusted values across all models and experiments with identical specifications. 
The key parameter settings are provided below:
{
\textbf{Qwen3-1.7B-Base \& Qwen3-4B-Base GRPO:}
\texttt{\\
adv\_estimator=grpo\\
train\_batch\_size=256\\
max\_prompt\_length=2,048\\
max\_response\_length=4,096\\
kl\_loss\_coef=0.001\\
kl\_loss\_type=low\_var\_kl\\
}
}
{\\
\textbf{Qwen3-1.7B-Base \& Qwen3-4B-Base DAPO:}
\texttt{\\
adv\_estimator=grpo\\
train\_batch\_size=256\\
max\_prompt\_length=2,048\\
max\_response\_length=4,096\\
overlong\_buffer.len=512\\
overlong\_buffer.penalty\_factor=1.0\\
loss\_agg\_mode=token-mean\\
}
}

\subsection{Text Normalization}
\label{app:normalization}
We conducted a rule-based text normalization for the rollouts. Specifically, we focus primarily on extracting mathematical expressions from both model responses and consensus nodes. We apply regular expression patterns to isolate critical numerical values and core equation relationships. The extraction process normalizes notational variations, including alternative multiplication symbols and fraction formats, and standardizes variable naming conventions to eliminate superficial discrepancies. This systematic normalization enhances both the accuracy and reliability of node matching.

\begin{table}[t]
\centering
\caption{Consistency metrics between Acc and NCR across models. }
\label{tab:leaderboard}
\begin{tabular}{lccc}
\toprule
Model            & Acc  & NCR  & TPN  \\
\midrule
Gemini-2.5-pro   & 91.8 & 80.1 & 299.42 \\
MiniMax-M2       & 93.3 & 67.9 & \textbf{60.61} \\
GPT-5.2          & \textbf{95.1} & \textbf{86.3} & 172.81 \\
DeepSeek-V4-pro  & 95.0 & 73.6 & 144.26 \\
\bottomrule
\end{tabular}
\end{table}

\section{Supplementary Results}
\label{app:results}
\subsection{Performance of Main LLMs}
\label{app:leaderboard}
We evaluated the performance of some main LLMs on our \DATANAME{}.
As shown in Table~\ref{tab:leaderboard}, although existing models achieve high accuracy in final answers (from 93.3\% to 95.1\%), their NCR scores vary significantly, spanning from 67.9\% to 86.3\%. This discrepancy indicates substantial room for improvement in the reasoning processes of current state-of-the-art large language models.
Additionally, TPN scores vary significantly across different models, indicating various reasoning strategies and efficiencies.

\begin{table}[t]
\centering
\caption{Human annotation validation metrics including annotation consistency and recall rate.}
\label{tab:human_validation}
\begin{tabular}{lcc}
\toprule
Metrics        &   Consistency & Recall Rate  \\
\midrule
Human Validation   & 62.2  & 90.5 \\
\bottomrule
\end{tabular}
\end{table}

\begin{table*}[t]
\centering
\caption{$\alpha$-sensitivity study on Qwen3-4B-Base.}
\label{tab:hyperpara}
\begin{tabular}{lcccccc}
\toprule
 & \textbf{AIME 24} & \textbf{AIME 25} & \textbf{GSM8K} & \textbf{MATH-500} & \multicolumn{2}{c}{\textbf{\DATANAME{}}} \\
$\alpha$ & Acc@16 & Acc@16 & Acc@1 & Acc@1 & Acc & NCR \\
\midrule
    0.0 & 36.7 & 33.3 & 67.1 & 54.0 & 65.0 & \textbf{79.8} \\
    0.1 & 40.0 & \textbf{43.3} & 90.7 & 61.2 & 84.4 & 77.1 \\
    0.2 & 40.0 & 36.7 & 92.0 & 58.6 & 84.8 & 78.1 \\
    0.3 & \textbf{43.3} & 33.3 & 91.2 & 58.8 & 84.6 & 76.5 \\
    0.4 & 40.0 & \textbf{43.3} & 91.1 & 62.2 & 85.4 & 77.8 \\
    0.5 & 33.3 & 36.7 & \textbf{92.8} & 60.6 & 84.9 & 78.4 \\
    0.6 & 30.0 & 26.7 & 92.5 & \textbf{65.8} & \textbf{86.8} & 78.4 \\
    0.7 & 36.7 & 23.3 & 91.3 & 60.6 & 84.3 & 76.8 \\
    0.8 & 30.0 & 30.0 & 91.7 & 60.0 & 84.5 & 75.3 \\
    0.9 & 33.3 & 33.3 & 91.1 & 60.0 & 84.3 & 76.6 \\
    1.0 & 26.7 & 26.7 & 92.3 & 58.4 & 84.1 & 77.4 \\
\bottomrule
\end{tabular}
\end{table*}

\subsection{Human Annotation Validation} \label{app:human_validation}
We conduct a human evaluation to validate the effectiveness of our method in identifying key nodes. Our setting and metrics are tailored to this task considering that the annotation data for consensus nodes consist of open-ended, free-form reasoning steps, and generic agreement measures such as Cohen's kappa coefficient have limited applicability to generative annotation. Extracting consensus nodes from multiple lengthy mathematical chains of thought also places substantial demands on the mathematical reasoning and language understanding of annotators. To ensure annotation quality and control experimental cost, we therefore evaluate a randomly sampled representative subset rather than the full dataset. Humans and LLMs differ in their underlying mechanisms for defining intermediate steps. Human annotators tend to retain only the most concise and critical content, whereas our LLM-based method aims to improve consensus and reward density by capturing finer-grained and more general steps. The average number of nodes extracted by humans is therefore necessarily smaller than that extracted by the LLM. Given this asymmetry, we report not only the exact agreement probability but also the recall of human-annotated nodes, defined as the proportion of human-identified key steps successfully captured by the LLM. Experiments show that although our automatic node extraction achieves 62.2\% agreement with human annotations, a recall of 90.5\% indicates that most discrepancies stem from humans not providing finer-grained annotations rather than from incorrect results. This further validates the effectiveness of our data collection method.

\subsection{Hyperparameter-sensitivity Study} \label{app:hyperpara}
To provide a comprehensive analysis of the model's sensitivity to $\alpha$, we conducted a new ablation study evaluating performance across $\alpha \in [0.0,1.0]$. The complete results are demonstrated in Table ~\ref{tab:hyperpara}.

\subsection{Performance of Qwen3-1.7B-Base}
We conduct our experiments on Qwen3-1.7B-Base as well for a scaling study and Table~\ref{tab:1.7b-results} demonstrates the results.
Experimental results indicate that the performance gains from our method are less pronounced for the 1.7B model compared to the 4B and 8B variants. We attribute this disparity to the limited emergent capabilities of smaller models. Integrating process rewards derived from formula nodes requires the model to simultaneously attend to both reasoning quality and result accuracy, a task that proves overly challenging for the 1.7B model. Consequently, it struggles to jointly improve both the process reward score (NCR) and the result reward score (Acc) during training.

\begin{table}[t]
\centering
\caption{Consistency metrics between Acc and NCR across models }
\label{tab:prm}
\begin{tabular}{lcc}
\toprule
Model            & GSM8K  & MATH500 \\
\midrule
$\text{Math-Shepherd}_\text{llama2-70B}$      & 93.2 & 44.5 \\
$\text{Math-Shepherd}_\text{llemma-34B}$      & 90.9 & 46.0 \\
$\text{Math-Shepherd}_\text{DeepSeek-67B}$    & 93.3 & 47.0 \\
$\text{AlphaMath}_\text{DeepSeekMath-Base-7B}$& 84.1 & 66.3 \\
$\text{OmegaPRM}_\text{Gemma2-27B}$           & 92.2 & 58.2 \\
\bottomrule
\end{tabular}
\vspace{-5mm}
\end{table}

\subsection{Comparison between ConsensusPR and PRM appraoches}
Although our approach provides a process-oriented reward, it differs significantly from traditional reinforcement learning algorithms built upon process reward models. 
In fact, our method can be viewed as a specialized form of outcome reward rather than a conventional process reward.
Nevertheless, we present several existing PRM methods in Table~\ref{tab:prm} for reference, including the scores of Math-Shepherd~\citep{DBLP:conf/acl/Math-Shepherd}, AlphaMath~\citep{AlphaMath}, and OmegaPRM~\citep{OmegaPRM} on GSM8K and MATH500. We directly cite data from prior studies, which reveal that these methods typically employ backbone models substantially larger than ours without achieving commensurate performance gains. This further demonstrates the superiority of our approach compared to standard PRM methods.

\end{document}